\documentclass{article}

\usepackage[final]{neurips_data_2022}

\usepackage[utf8]{inputenc} 
\usepackage[T1]{fontenc}    
\usepackage{hyperref}       
\usepackage{url}            
\usepackage{booktabs}       
\usepackage{amsfonts}       
\usepackage{amsmath}
\usepackage{nicefrac}       
\usepackage{microtype}      
\usepackage{xcolor}         
\usepackage{graphicx}
\usepackage{float}
\usepackage{listings}
\usepackage{xcolor}
\usepackage{hyperref}
\usepackage{caption}
\usepackage{subcaption}
\usepackage{bold-extra}
\usepackage{tablefootnote} 
\usepackage{booktabs}
\usepackage[inline]{enumitem}
\usepackage{courier}

\hypersetup{
  colorlinks   = true, 
  urlcolor     = blue, 
  linkcolor    = blue, 
  citecolor   = blue 
}

\newcommand{\wikid}[1]{\href{https://www.wikidata.org/wiki/#1}{#1}}
\newcommand{\ner}[2]{[\textbf{#1}]\textsubscript{[#2]}}

\newif\ifreviewer

\reviewertrue

\ifreviewer 
\newcommand{\meta}[1]{\textcolor{brown}{$_{meta}$[#1]}}
\newcommand{\aman}[1]{\textcolor{cyan}{$_{aman}$[#1]}}
\newcommand{\raheleh}[1]{\textcolor{red}{$_{raheleh}$[#1]}}
\newcommand{\sneha}[1]{\textcolor{blue}{$_{sneha}$[#1]}}
\newcommand{\shubh}[1]{\textcolor{magenta}{$_{shubh}$[#1]}}

\else
\newcommand{\meta}[1]{{}}
\newcommand{\aman}[1]{{}}
\newcommand{\raheleh}[1]{{}}
\newcommand{\sneha}[1]{{}}
\newcommand{\shubh}[1]{{}}

\fi

\newcommand{\definition}[2]{\textbf{#1}: #2}
\newcommand{\TweetNERD}{\texttt{TweetNERD}}
\newcommand{\TweetNERDOOD}{\TweetNERD{}-\texttt{OOD}}
\newcommand{\TweetNERDAcademic}{\TweetNERD{}-\texttt{Academic}}

\title{\TweetNERD{} - End to End Entity Linking Benchmark for Tweets}

\author{%
    Shubhanshu Mishra\thanks{Corresponding Author}\\
    \And
    Aman Saini\\
    \And
    Raheleh Makki\\
    \And
    Sneha Mehta\\
    \And
    Aria Haghighi\\
    \And
    Ali Mollahosseini\\
    Twitter, Inc.\\
    \texttt{\{smishra,amansaini,rmakki,snehamehta\}@twitter.com}\\
    \texttt{\{ahaghighi,amollahosseini\}@twitter.com}\\
}

\definecolor{codegreen}{rgb}{0,0.6,0}
\definecolor{codegray}{rgb}{0.5,0.5,0.5}
\definecolor{codepurple}{rgb}{0.58,0,0.82}
\definecolor{backcolour}{rgb}{1,1,1}

\lstdefinestyle{mystyle}{
    backgroundcolor=\color{backcolour},   
    commentstyle=\color{codegreen},
    keywordstyle=\color{magenta},
    numberstyle=\tiny\color{codegray},
    stringstyle=\color{codepurple},
    basicstyle=\ttfamily,
    breakatwhitespace=false,         
    breaklines=true,                 
    captionpos=b,                    
    keepspaces=true,                 
    numbers=left,                    
    numbersep=5pt,                  
    showspaces=false,                
    showstringspaces=false,
    showtabs=false,                  
    tabsize=2
}

\begin{document}

\maketitle

\begin{abstract}
Named Entity Recognition and Disambiguation (NERD) systems are foundational for information retrieval, question answering, event detection, and other natural language processing (NLP) applications. We introduce \TweetNERD{}, a dataset of 340K+ Tweets across 2010-2021, for benchmarking NERD systems on Tweets. This is the largest and most temporally diverse open sourced dataset benchmark for NERD on Tweets and can be used to facilitate research in this area. We describe evaluation setup with \TweetNERD{} for three NERD tasks: Named Entity Recognition (NER), Entity Linking with True Spans (EL), and End to End Entity Linking (End2End); and provide performance of existing publicly available methods on specific \TweetNERD{} splits. \TweetNERD{} is available at: \url{https://doi.org/10.5281/zenodo.6617192} under Creative Commons Attribution 4.0 International (CC BY 4.0) license \citep{mishra_shubhanshu_2022_6617192}. Check out more details at \url{https://github.com/twitter-research/TweetNERD}.

\end{abstract}

\section{Introduction}

\begin{figure}[!htbp]
    \centering
    \includegraphics[width=0.85\linewidth]{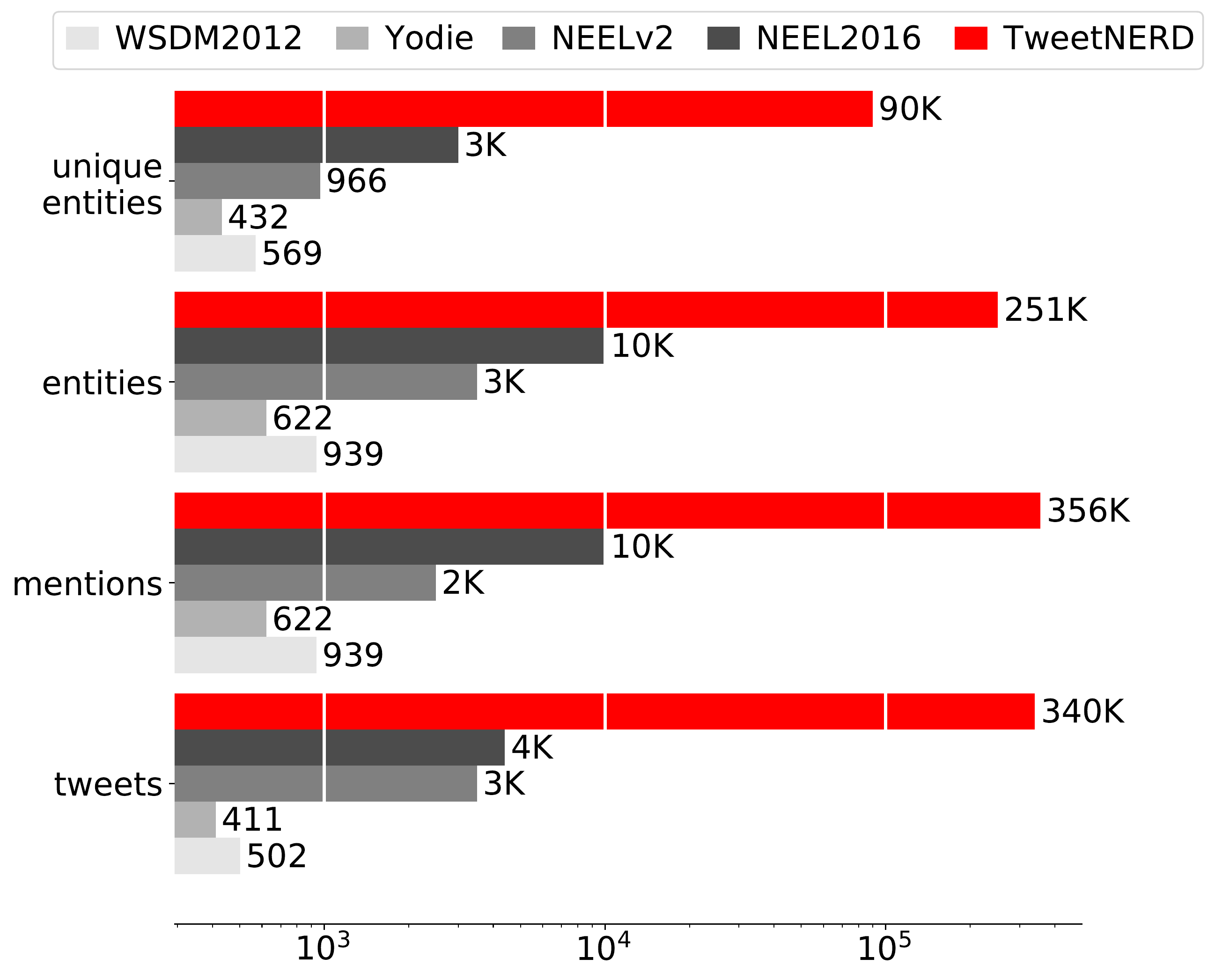}
    \caption{Comparison with existing Tweet entity linking datasets}
    \label{fig:data_stats}
\end{figure}

Named Entity Recognition and Disambiguation (NERD) \citep{WkifyMihalcea2007,cucerzan-2007-large,derczynski_analysis_2015,kulkarni_collective_2009} is the task of identifying important mentions or Named Entities in the text and linking those mentions to corresponding entities in an underlying Knowledge Base (KB). The KB can be any public knowledge repository like Wikipedia or a custom knowledge graph specific to the domain. 
NERD for social media text \citep{derczynski_analysis_2015,mishra-diesner-2016-semi,mishra-mdmt-2019}, in particular Tweets is challenging because of the short textual context owing to the 280 character limit of Tweets. There exist multiple datasets \citep{derczynski_analysis_2015,mishra-mdmt-2019, dredze-etal-2016-twitter, derczynski-etal-2016-broad, spina2012corpus, NEEL2016, yang-chang-2015-mart, fang-chang-2014-entity, locke2009named, Meij-WSDM2012, gorrell_using_2015} for developing and evaluating NERD methods on Tweets. However, these datasets have limited set of Tweets, are temporally biased (i.e. Tweets are from a short time period, more details in section \ref{sec:appendix_academic_data_details}), or are no longer valid because of deleted Tweets (see Table \ref{tab:academic-split}). In this work, we introduce a new dataset called \TweetNERD{} which consists of 340K+ Tweets annotated with entity mentions and linked to entities in Wikidata (a large scale multilingual publicly editable KB). \TweetNERD{} addresses the issues in existing NERD datasets for Tweets by including Tweets from a broader time window, applying consistent annotations, and including the largest collection of annotated Tweets for NERD tasks. Figure \ref{fig:data_stats} compares \TweetNERD{} with existing Tweet entity linking datasets, proving its increases scale. Furthermore, we describe two splits of the dataset which we use for evaluation. These splits called \TweetNERDOOD{} and \TweetNERDAcademic{} allow assessing out of domain (OOD) and temporal generalization respectively. \TweetNERDOOD{} split consists of Tweets in a shorter time frame that are over-sampled with harder to disambiguate entities. It is useful to assess out of domain performance. Conversely, \TweetNERDAcademic{} split is a temporally diverse dataset of non-deleted Tweets from a collection of existing academic benchmarks that have been re-annotated with the new annotation guidelines. \TweetNERD{} has already been used by \citet{hebert-etal-2022-robust} for evaluating dense retrieval for candidate generation in presence of noisy NER spans. \TweetNERD{} should also foster research in better utilization of social graph context of Tweets \citep{kulkarni-etal-2021-lmsoc-approach,li-etal-2022-ntulm} in improving NERD task performance, and assessment of bias in NERD systems \citep{NERBias}. \TweetNERD{} is available at: \url{https://doi.org/10.5281/zenodo.6617192} under Creative Commons Attribution 4.0 International (CC BY 4.0) license \citep{mishra_shubhanshu_2022_6617192}. Check out more details at \url{https://github.com/twitter-research/TweetNERD}.

\subsection{Related works}
Named Entity Recognition and Disambiguation (NERD) is a prominent information extraction task. There exist multiple datasets \citep{derczynski_analysis_2015,mishra-mdmt-2019, dredze-etal-2016-twitter, derczynski-etal-2016-broad, spina2012corpus, NEEL2016, yang-chang-2015-mart, fang-chang-2014-entity, locke2009named, Meij-WSDM2012, gorrell_using_2015} for either doing Named Entity Recognition (NER), NERD, Cross Domain Co-reference Retreival (CDCR), or Entity Relevance. Most datasets were created by sampling Tweets from a given time-period and then annotating them either for NER alone or for NERD. The annotation also differs by linking to either DBPedia \citep{gorrell_using_2015}, Wikipedia \citep{NEEL2016}, or Freebase \citep{fang-chang-2014-entity} as the possible knowledge base. Our work closely follows the annotation process of \citet{gorrell_using_2015} of linking entities using a crowd sourcing platform and doing both NER and Entity Disambiguation tasks. Our data collection process differs in terms of sampling Tweets from a diverse temporal window and the inclusion of more diverse set of entities (see section \ref{sec:sampling}).

\section{Terminology} We use the following terminology throughout the rest of the paper: 
\begin{enumerate*}[label=(\arabic*)]
\item \definition{knowledge base (\texttt{KB})}{Underlying knowledge base of entities, we use Wikidata \citep{vrandecic_wikidata_2014}.}
\item \definition{document id ($id$)}{Id of the document with entities, and optional meta-data e.g. date;}
\item \definition{mention ($m$)}{a phrase in document $d$ identified by a start offset \textbf{$s$} and end offset \textbf{$e$};}
\item \definition{start ($s$)}{starting offset of mention $m$. The offset is dependent on the encoding of the data (\TweetNERD{} uses byte offsets for the text encoded using \texttt{utf-16-be});}
\item \definition{end ($e$)}{ending offset of mention $m$ in the same format as ($s$), such that $len(m) = e-s$;}
\item \definition{\texttt{NIL}}{If a mention can't be linked to any entity in \texttt{KB};}
\item \definition{entity id ($eid$)}{Linked entity Id in \texttt{KB} or \texttt{NIL}; and}
\item \definition{candidate set ($C$)}{Possible candidates for $m$ in \texttt{KB} and \texttt{NIL}.}
\end{enumerate*}

\section{Annotation Setup}
\label{sec:annotation_setup}
\paragraph{Annotators} We leveraged a team of trained in-house annotators who utilized a custom annotation interface to annotate the Tweets. A pool of annotators was trained with detailed labeling guidelines and multiple rounds of training iterations before actually starting to annotate the Tweets in \TweetNERD{}. The guidelines included examples of Tweets with linked entities, and instructions on how to disambiguate between potential candidates using the Tweet context, media, time and other factors. A much simplified version of the interface is shown for the purpose of illustration (see Figure \ref{fig:interface}). The annotators were required to pass a qualification quiz demonstrating their understanding of the task to be eligible as an annotator.
\paragraph{Annotation Task} The annotation task required identifying all mentions $m$ in a Tweet and assigning a Wikidata ID, $eid$, for each $m$. 
The annotators had to highlight the mention and then use Wikidata search interface to find the correct $eid$ (e.g. $m$=\texttt{Twitter} and $eid$=\texttt{\href{https://www.wikidata.org/wiki/Q918}{Q918}}).
Annotators could edit the search phrase to differ from $m$ to correct for spelling errors, or expand it with additional words in order to find a suitable entity.
If there is no valid Wikidata ID for $m$, annotators assigned $eid$=\texttt{NOT FOUND}. If annotators thought that the Tweet context is not clear enough to disambiguate between the returned candidates they assigned $eid$=\texttt{AMBIGUOUS}. The Wikidata ID for a given Wikipedia page is obtained by clicking on the Wikidata item link located on the left panel of the Wikipedia page. \TweetNERD{} annotation was done in batches where around 25K Tweet ids for each batch were sampled via the setup described in the next section. We annotated a total of 14 batches for the \TweetNERD{} dataset.

\paragraph{Eligible Mentions} Annotators were instructed to select mentions $m$ in a Tweet which refer to the longest phrase corresponding to a named entity that can be identified as a Person, Organization, Place etc. (see Table \ref{tab:hcomp_types} for full list and details). A mention can also be contained within a hashtag if it corresponds to an entity e.g. \texttt{\#FIFA}. 

\paragraph{Correct Candidate} Annotators were instructed to prefer an $eid$ which is likely to have a Wikipedia page. The most appropriate $eid$ could depend on the following:
\begin{enumerate*}[label=(\alph*)]
    \item full text of the Tweet, 
    \item the URL or media attached to the Tweet, 
    \item the temporal context of the Tweet (annotators can search for $m$ on Twitter around the same date as the Tweet), 
    \item the Tweet thread it is part of (i.e. which Tweet it is replying to and the list of Tweets which replied to it)
    \item the user of the Tweet
\end{enumerate*}.

\paragraph{Annotation Aggregation} Each Tweet was annotated by \textit{three} annotators and $(m, eid)$ pairs that were selected by \textit{ at least two} annotators were considered \textbf{gold annotations}. We include all annotations (including non-gold) as part of the final dataset to support additional analysis (e.g. studying annotation noise). 

\begin{figure}[!htbp]
    \centering
    \begin{tabular}{p{0.9\linewidth}}
    \toprule
    \textbf{Id=1}: I love \ner{Twitter}{ENTITY}\newline
    {
    Candidates: \textbf{\wikid{Q918}}, {NOT FOUND}, {AMBIGUOUS}
    }
    \\
    \midrule
    \textbf{Id=2}: \ner{Paris}{ENTITY} is regarded as the world's fashion capital\newline
    {Candidates: \textbf{\wikid{Q90}}, \wikid{Q79917}, {NOT FOUND}, {AMBIGUOUS}}\\
    \midrule
    \textbf{Id=3}: \ner{Anil}{ENTITY} is playing\newline
    { Candidates: {NOT FOUND}, \textbf{AMBIGUOUS}}\\
    \bottomrule
    \end{tabular}
    \caption{\textbf{Simplified version of the annotation interface.} Selected mentions and entities are in \textbf{Bold}. Important thing to note is that the annotators are shown only the Tweet text. They use the functionality provided in the interface to query the eligible knowledge base candidates. Each annotator can select multiple mentions in a Tweet but link each mention ($m$) to only a single Entity Id ($eid$).}
    \label{fig:interface}
\end{figure}

\paragraph{Difficulty of the Annotation Task}
\label{sec:task_difficulty}
Entity Linking is inherently a difficult task due to name variations (multiple surface forms for the same entity) and entity ambiguity (multiple entities for the same mention)~\citep{shen2014entity}. In addition, based on the type of application and the coverage of the underlying knowledge base this task can become challenging even for humans. E.g. we asked the annotators to link a mention to the most specific entity in the knowledge base (i.e. Wikidata), this assumption forces all other candidate entities (even if close) for that mention as incorrect. For instance, if a Tweet is about the Academy Awards this year (2022), we only consider \wikid{Q66707597} (94th Academy Awards) as the correct entity and not \wikid{Q19020} (Academy Awards), while \wikid{Q19020} is the correct entity if the Tweet is about Academy Awards in general. While this allows for temporally sensitive annotations, it makes the task difficult compared to most classification tasks, leading to a negative impact on inter-annotator agreement (see discussion in section \ref{sec:inter-annotator-agreement}). 

\begin{table*}[]
\centering
\caption{Example of types of entities to identify in the text}
    \label{tab:hcomp_types}
\begin{tabular}{@{}p{0.2\linewidth}p{0.75\linewidth}@{}}
\toprule
\textbf{Type}       & \textbf{Examples}                                                                                                                                                                                                                                                                 \\ \midrule
Person              & \begin{tabular}[c]{@{}p{\linewidth}@{}}Politicians, sports players, artists, celebrities, fictional characters, scientists, singers, musicians, journalists, social media celebrities, and others\\ Examples: Kanye West, Sachin Tendulkar, Donald Trump, Harry Potter, Jon Snow\end{tabular} \\
\midrule
Place               & \begin{tabular}[c]{@{}l@{}}Countries, Cities, Monuments, Parks, rivers, and others\\ Examples: Paris, Nigeria, Statue of Liberty\end{tabular}                                                                                                                                     \\
\midrule
Organization        & \begin{tabular}[c]{@{}p{\linewidth}@{}}Companies, governments, NGOs, social movements, music bands, sports teams, social organizations, volunteer organizations, and others\\ Examples: Backstreet Boys, Los Angeles Lakers, Black Lives Matter\end{tabular}                                  \\
\midrule
Products            & \begin{tabular}[c]{@{}p{\linewidth}@{}}Websites, Softwares, applications, video games, technology gadgets, devices, and others\\ Examples:  PlayStation, iPhone, GoFundMe, Roblox\end{tabular}                                                                                                \\
\midrule
Works of Art        & \begin{tabular}[c]{@{}p{\linewidth}@{}}Movies, Albums, Books, Comics, Video Games, TV Shows, Social Media videos, and others\\ Examples: Friends, The Office, Lupin\end{tabular}                                                                                                              \\
\midrule
Scientific Concepts & \begin{tabular}[c]{@{}p{\linewidth}@{}}Names of diseases, drugs, names of algorithms, scientific methods and techniques, scientific names of organisms, names of disasters, and others\\ Examples: COVID-19, SARS-COV19, Hurricane Katrina, Cyclone Idai\end{tabular}                         \\ 
\bottomrule
\end{tabular}

\end{table*}

\section{Tweet End To End Entity Linking Dataset}
\subsection{Sampling}
\label{sec:sampling}

\TweetNERD{} consists of English Tweets most of which were created between Jan 2020 and Dec 2021. Tweet language was identified using the Twitter Public API endpoint. Additionally, we discarded Tweets which were NSFW\footnote{NSFW - Not Safe For Work}, too short ($\le 10$ space separated tokens), and included $\ge 2$ URLs or $\ge 2$ user mentions or $\ge 3$ Hashtags. 
Since the dataset was annotated in batches, we were able to improve our sampling technique with each batch. Our initial approach of upsampling Tweets with high retweets and likes (Tweet-actions) resulted in a large proportion of Tweets with empty annotations. To mitigate this, we experimented with approaches which select Tweets that are more likely to have an entity. Some of these approaches included:
\begin{enumerate*}[label=(\alph*)]
    \item using in-house NER models \citep{NERBias,eskander_ramy_2020_7014432} to check for NER mentions,
    \item using phrase matching techniques \citep{mishra-diesner-2016-semi} to match phrases from Tweet text with the Wikidata entity titles,
    \item sampling based on phrase entropy to detect difficult phrases (described in the next paragraph),
    \item overall Tweet favorite based sampling, and
    \item search page click based sampling
\end{enumerate*}. 
Within each approach, we perform a stratified sampling to select Tweets equally from each sampling bucket.
The full dataset \TweetNERD{} is comprised of different proportions of each of these buckets.

\paragraph{Entropy based sampling}
\label{para:entropy-sampling}
We wanted to include tweets containing phrases representing a diverse set of wikidata entities in terms of entity popularity as well as disambiguation difficulty. We used aggregate wikipedia page views ($eid_{views}$) across all language pages of a wikidata entity as a proxy for its popularity. Then the phrase entropy was defined as $H = \sum p*log(p)$ using the probability $p = p(eid | m) = eid_{views} / \sum{eid_{views}}$. Each phrase is then classified as one of the high, medium, or low entropy phrase using the entropy score distribution. Finally, we sample an equal number of Tweets from each phrase entropy bucket. 

\subsection{Data Splits}
\label{sec:data-splits}

While \TweetNERD{} consistes of 340K+ Tweets, we highlight two explicit data splits of \TweetNERD{}, namely \TweetNERDOOD{} and \TweetNERDAcademic{}, which have been used as test sets for evaluation in this paper. The purpose of these two splits is to measure out of domain performance and temporal generalization respectively.

\paragraph{\TweetNERDOOD{}} It is a subset of 25K Tweets used for evaluating existing named entity recognition and linking models. \TweetNERDOOD{} is sampled in equal proportion based on the entropy of the contained NER mentions. Mentions with few, less diverse candidates fall in the low entropy buckets whereas mentions with many, high diversity candidates fall into the high entropy buckets. We first sample Tweets into high, medium and low entropy mention buckets, and then perform stratified sampling based on Tweet actions to divide these buckets into sub-buckets. This approach helps us to evaluate all models against a variety of Tweets with varying levels of difficulty and popularity.

\paragraph{\TweetNERDAcademic{}} It is a subset of 30K Tweets to benchmark entity linking systems on Tweets already sampled in existing academic benchmarks (mostly from \citep{derczynski_analysis_2015,mishra-mdmt-2019}). We identify all the Tweet ids across existing NERD, NER, NED, and syntatic NLP task datasets for Tweets and hydrate these ids using the Public Twitter API. We ended up with 30,119 Tweets across these datasets which are still available (see Table \ref{tab:academic-split}). Its important to note that these Tweets were annotated again using our latest annotation setup to comply with the \TweetNERD{}  guidelines. Our intention for including this split is to add a layer of temporally diverse and already benchmarked datasets. 

\paragraph{Re-annotation of academic benchmarks in \TweetNERDAcademic{}} We re-annotate the academic benchmark datasets in \TweetNERDAcademic{} using our guidelines and setup to ensure consistency of these annotations with the rest of our dataset. This choice was made as opposed to including the existing annotations from these datasets for the following reasons. First, not all of these datasets are annotated for the end to end NERD task, i.e. some only have NER and some only have NED annotations. Second, the knowledge base used for each NERD annotation is not Wikidata. Instead, some datasets link to DBpedia, some to English Wikipedia. Third, the notion of entities to annotate varies across the datasets and would require a lot of reconciliation to make a consistent benchmark dataset, e.g. \citet{NEEL2016} annotates Hashtags and user mentions as entities but \TweetNERD{} does not allow mentions to be tagged as entities. Finally, many of the Tweets (20-40\%, see table \ref{tab:academic-split}) from these datasets are not available via the public API, however, those which are still available are likely to be available for a longer duration which makes this benchmark more stable. We show some examples of annotations in \TweetNERDAcademic{} versus existing academic benchmarks in table \ref{tab:annotation-compare}. Detailed description of each of these datasets is provided in section \ref{sec:appendix_academic_data_details}. Finally, we observed high overlap between \TweetNERDAcademic{} and academic datasets. E.g. using Yodie as the closest academic dataset in terms of our annotation guidelines, we found that \TweetNERDAcademic{} matches 77\% Yodie mention level annotations as well as 87\% mention annotations at the Tweet level. At the mention-entity level \TweetNERDAcademic{} matches 65\% Yodie annotations and 80\% at the Tweet level (we map DBPedia entity annotations in Yodie to their Wikidata ID).

\begin{table}
\caption{Annotations in \TweetNERDAcademic{} versus annotations in existing benchmarks.}
\label{tab:annotation-compare}
\begin{tabular}{p{\linewidth}}
\toprule
\textbf{Text}: Press release:"Will England fans be hit by penalties on their next energy bill?" Please make it stop. \textbf{Yodie}: England (\href{http://dbpedia.org/resource/England}{Dbp:England}); \textbf{\TweetNERD{}}: England (\wikid{Q21}) \\ 
\midrule
\textbf{Text}: \#DMG \#GILDEMEISTER presents the new GILDEMEISTER energy monitor, read more at [URL]. \textbf{Yodie}: GILDEMEISTER (6, 18, \href{http://dbpedia.org/resource/Gildemeister_AG}{Dbp:Gildemeister\_AG}), GILDEMEISTER (36, 48, \href{http://dbpedia.org/resource/Gildemeister_AG}{Dbp:Gildemeister\_AG}); \textbf{\TweetNERD{}}: GILDEMEISTER (6, 18, \wikid{Q100151808}), GILDEMEISTER (36, 48, \wikid{Q100151808}) \\
\midrule
\textbf{Text}: Wiz Khalifa went suit shopping with Max Headroom. \#grammys \#80s [URL]. \textbf{TGX}: Max Headroom (NA, NA, NA); \textbf{\TweetNERD{}}: Wiz Khalifa (0, 11, \wikid{Q117139}), Max Headroom (36, 48, \wikid{Q1912691}) \\
\bottomrule
\end{tabular}
\end{table}

\paragraph{Flexibility for Further Analysis} As seen above, we have identified two subsets of the dataset (\TweetNERDOOD{} and \TweetNERDAcademic{}) which we use as test sets for evaluation in this paper. While these two datasets can be used for standard benchmarking for tasks similar to those presented in this paper, we would like to emphasize the flexibility of \TweetNERD{} in evaluating a wide range of tasks. For example, one could split the full \TweetNERD{} dataset temporally to test existing models for temporal generalization or one could split \TweetNERD{} based on seen and unseen mentions and entities to assess robustness. \TweetNERD{} can also be randomly split into train, validation, and test splits that can be used to evaluate in-domain performance of models. To align ourselves with the traditional machine learning benchmark formats, we also provide canonical train, validation, and test splits of the data created by extracting random samples of 25K tweets for test and 5K for validation from \TweetNERD{} excluding \TweetNERDOOD{} and \TweetNERDAcademic{}. While we do not report any results on this test split in this paper, we encourage researchers to use these splits along with \TweetNERDOOD{} and \TweetNERDAcademic{} to ensure reproducibility.

\paragraph{Adapting to Temporal Dynamics of Knowledge Bases}
Knowledge Bases are dynamic and new entities are added with time and since NERD datasets are not updated with time there might be discrepancies in model evaluation with reference to a static NERD test set. This is a common limitation of Entity linking evaluation. In \TweetNERD{} this would only affect the NIL predictions as opposed to linking predictions. An entity which in 2014 was marked as NIL (because of absence from Wikidata) may be marked correctly now.
This can be addressed easily by factoring in the creation date of the entity in Wikidata. This way any entity whose creation date in Wikidata is after the Tweet date can be marked as NIL. This can allow for temporal evaluation.

\subsection{Data Statistics}
\label{sec:data-stats}

\begin{table}[]
\centering
\caption{Details of \TweetNERDAcademic{} (same Tweet could occur in multiple datasets).}
\label{tab:academic-split}
    \begin{tabular}{llrrr}
    \toprule
    \textbf{dataset} & \textbf{Tasks}                          & \textbf{Total Tweets} &  \textbf{Found Tweets} & \textbf{Found \%}\\
    \midrule
    \textbf{Tgx} \citep{dredze-etal-2016-twitter}	&	CDCR	&	15,313	&	9,790	&	63.9	\\
\textbf{Broad}  \citep{derczynski-etal-2016-broad}	&	NER	&	8,633	&	6,913	&	80.1	\\
\textbf{Entity Profiling} \citep{spina2012corpus}	&	NER	&	9,235	&	6,352	&	68.8	\\
\textbf{NEEL 2016}  \citep{NEEL2016}	&	NERD	&	9,289	&	2,336	&	25.1	\\
\textbf{NEEL v2}   \citep{yang-chang-2015-mart}	&	NERD	&	3,503	&	2,089	&	59.6	\\
\textbf{\citet{fang-chang-2014-entity}}	&	NERD	&	2,419	&	1,662	&	68.7	\\
\textbf{Twitter NEED}    \citep{locke2009named}	&	NERD \& IR	&	2,501	&	1,549	&	61.9	\\
\textbf{Ark POS}     \citep{gimpel-etal-2011-part}	&	POS	&	2,374	&	1,313	&	55.3	\\
\textbf{WikiD}	&	NED	&	1,000	&	504	&	50.4	\\
\textbf{WSDM2012}  \citep{Meij-WSDM2012}	&	Relevance	&	502	&	415	&	82.7	\\
\textbf{Yodie} \citep{gorrell_using_2015}	&	NERD	&	411	&	288	&	70.1	\\
    \bottomrule
    \end{tabular}
\end{table}

\begin{figure}[!htbp]
\centering
\includegraphics[width=0.6\linewidth]{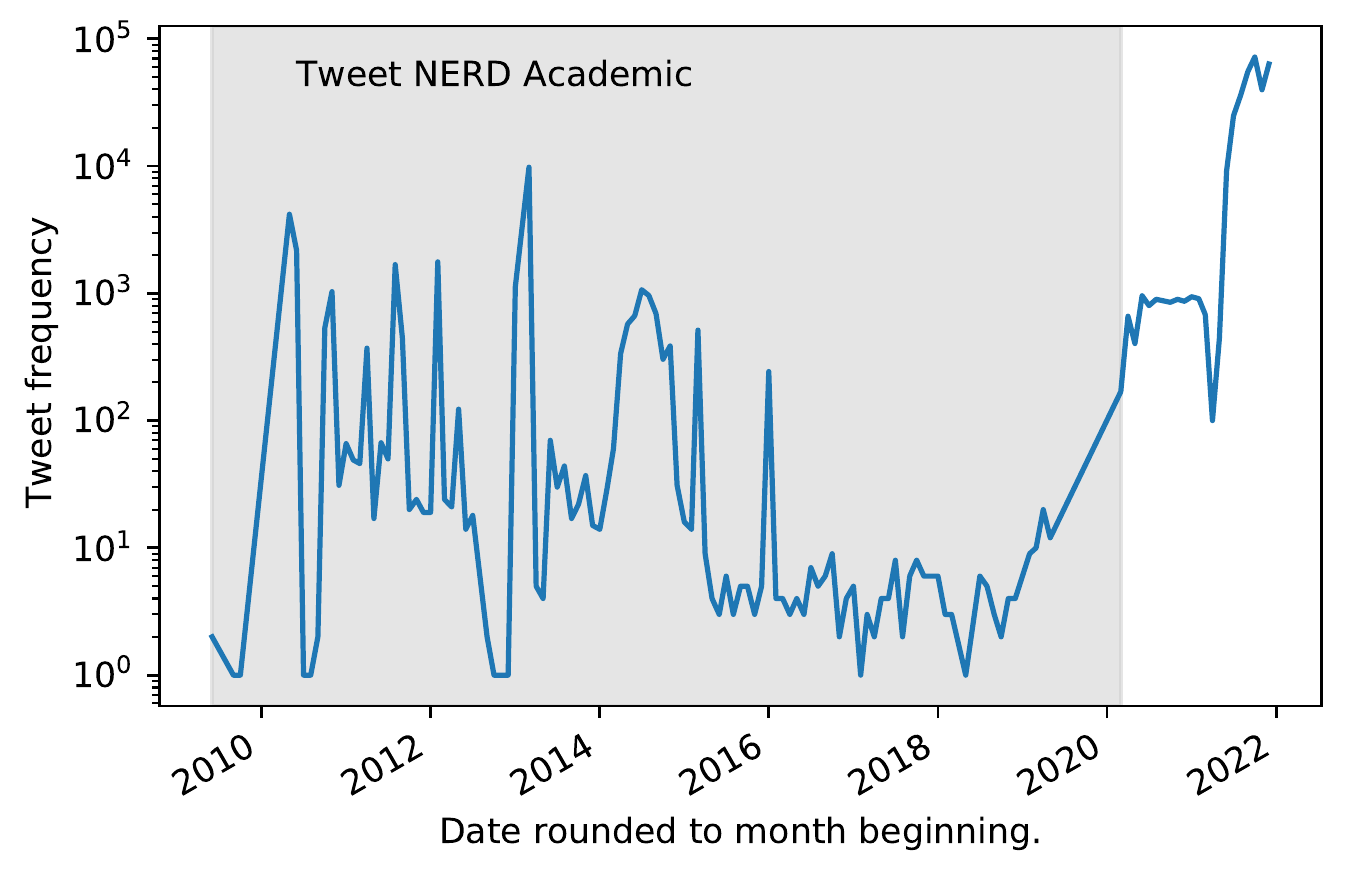}
\caption{Temporal Frequency of Tweets in the \TweetNERD. Time period of \TweetNERDAcademic{} highlighted in Grey.}
\label{fig:data_temporal_freq}    
\end{figure}

\begin{table*}[]
    \centering
    \caption{Salient entities, mentions, and mention-entity pairs in \TweetNERD{} full dataset and subset. Entity refers to $eid$ - the linked Wikidata ID, Mention refers to $m$ - the annotated phrase in the Tweet, and Mention-Entity refers to $(m, eid)$ - a unique tuple of <mention, entity>. }
    \label{tab:salient_entities}
    \begin{tabular}{p{0.95\linewidth}}
        \toprule
        \textbf{Full data set} \\
        \midrule
         \textbf{Mention Entity}: Total: 356345, Unique: 166379 \\
Head: "'grammys' <Q630124>" (6272), "'mark lee' <Q26689986>" (2341), "'aria' <AMBIGUOUS>" (2103), "'whatsapp' <Q1049511>" (1521), "'isabella' <AMBIGUOUS>" (1260) \\
Mid: "'david mabuza' <Q1174142>" (2), "'neha sharma' <Q863745>" (2) \\
Tail: "'ian darke' <Q5981359>" (1), "'antony perumbavoor' <Q55604079>" (1), "'sansone' <NOT FOUND>" (1), "'prairie state college' <NOT FOUND>" (1), "'konga' <NOT FOUND>" (1)  \\
    \midrule
    \textbf{Mention}: Total: 356345, Unique: 143762 \\
    Head: 'grammys' (7059), 'aria' (2461), 'mark lee' (2342), 'whatsapp' (1602), 'isabella' (1471) \\
    Mid: 'nam joo hyuk' (2), 'sharpsburg' (2) \\
    Tail: 'iain banks' (1), 'michael odewale' (1), 'chlorine cougs' (1), 'rock your baby' (1), 'georgia dome' (1) \\
    \midrule
    \textbf{Entity}: Total: 356345, Unique: 90938 \\
    Head: 'NOT FOUND' (59704), 'AMBIGUOUS' (44752), 'Q630124' (7886), 'Q26689986' (2364), 'Q108112350' (2094) \\
    Mid: 'Q9196194' (2), 'Q331613' (2) \\
    Tail: 'Q107362802' (1), 'Q81101633' (1), 'Q17361809' (1), 'Q1177' (1), 'Q741395' (1) \\
    \midrule
    \textbf{} \\
    \midrule
    \textbf{Without \TweetNERDAcademic} \\
    \midrule
    \textbf{Mention Entity}: Total: 312581, Unique: 159468 \\
Head: "'mark lee' <Q26689986>" (2341), "'aria' <AMBIGUOUS>" (2103), "'whatsapp' <Q1049511>" (1521), "'isabella' <AMBIGUOUS>" (1260), "'tajin' <Q3376620>" (1016) \\
Mid: "'cannes2021' <Q42369>" (2), "'zeynep' <NOT FOUND>" (2) \\
Tail: "'slave play' <Q69387965>" (1), "'Prada' <Q193136>" (1), "'gansu' <Q42392>" (1), "'iowa state capitol' <Q2977124>" (1), "'konga' <NOT FOUND>" (1) \\
    \midrule
    \textbf{Mention}: Total: 312581, Unique: 137782 \\
Head: 'aria' (2461), 'mark lee' (2342), 'whatsapp' (1602), 'isabella' (1471), 'matilda' (1434) \\
Mid: 'jamelia' (2), 'mohammad rafi' (2) \\
Tail: 'petr yan' (1), 'wooiyik' (1), 'billie dove' (1), 'bucks fizz' (1), 'georgia dome' (1) \\
    \midrule
    \textbf{Entity}: Total: 312581, Unique: 87430 \\
    Head: 'NOT FOUND' (58678), 'AMBIGUOUS' (44202), 'Q26689986' (2364), 'Q108112350' (2094), 'Q1049511' (1554) \\
    Mid: 'Q1186977' (2), 'Q983026' (2) \\
    Tail: 'Q455833' (1), 'Q3283342' (1), 'Q17183770' (1), 'Q7491877' (1), 'Q30308127' (1) \\
    \bottomrule
    \end{tabular}
    
\end{table*}

\paragraph{\TweetNERD.} \TweetNERD{} consists of 340K unique Tweets that collectively contain a total of 356K mentions that are linked to 90K unique entities. Of the 356K mentions, 251K are linked to non-NIL entities, and 104K to NIL entities. As can be observed in Figure \ref{fig:data_stats}, \TweetNERD{} is the largest data set compared to existing benchmark datasets for Tweet entity linking. More details about the salient mentions, entities, and mention-entity pairs in \TweetNERD{} can be found in Table \ref{tab:salient_entities}.

\paragraph{Temporal Distribution of Dataset.}
\label{sec:temporal-dist}

Our dataset consists of 340K Tweets spread across a period of 12 years from 2010 to 2021. This includes a smaller but temporally diverse subset which includes Tweets from existing academic benchmarks, re-annotated using our guidelines. If we remove the academic benchmarks, the resulting dataset consists of 310K Tweets spread from 2020-01 to 2021-12. \TweetNERD{} includes a non uniform sampling across time.

\subsection{Inter-annotator agreement}
\label{sec:inter-annotator-agreement}
\paragraph{Limitations of current inter-annotator agreement measures for NERD tasks} All Tweets in \TweetNERD{} are annotated by three annotators. For classification tasks Cohen’s Kappa \citep{McHugh2012-yr} is considerrd a standard measure of inter-annotator agreement (IAA). However, for NERD tasks, Kappa is not the most relevant measure, as noted in multiple studies (\cite{hripcsak2005agreement, grouin2011proposal}). The main issue with Kappa is its requirement of negative classes which is not known for NER and NERD tasks. Furthermore, NERD task involves a sequence of words or in our case offsets in text making the number of items variable for each text. A workaround is to use Kappa at token level. However, this results in additional issues. First, annotations are done at the Tweet level instead of token level and for our task tokens will depend on the tokenizer used. Second, token level annotaiton lead to an abundance of "O" tags for NER which will overwhelm the kappa statistics. In \citet{derczynski-etal-2016-broad} the evaluation is done using F1 measure between annotations of two annotators. This is reasonable when we have a fixed set of annotators doing annotation on all the Tweets. However, this is not possible for \TweetNERD{} as the annotations were collected from a crowd sourcing system where different set of annotators may annotate different Tweets. Hence, the only approach for calculating agreement in our case is agreement among annotators. 

\paragraph{\TweetNERD{} NERD agreement} We compute inter-annotator agreement at \textbf{mention $m$ and mention-entity $(m, eid)$} levels.
69\% mentions have a majority ($\ge 2$) agreement, of which 38\% have agreement from all three annotators.
17\% of mention-entities have 100\% agreement across all three annotators, 41\% have majority ($\ge 2$), and 59\% have only single annotator. 40\% mention-entities in \TweetNERDOOD{} and 57\% in \TweetNERDAcademic{} have majority agreement.
If we consolidate \texttt{AMBIGOUS} and \texttt{NOT FOUND} $eid$ as \texttt{NIL} the majority agreement goes up to 47\%.
At the Tweet level, 30\% Tweets have majority agreement across all annotated mention-entities.
These agreement scores highlight the difficulty and ambiguity of the end to end entity linking annotation task as described in Section \ref{sec:task_difficulty}.
While it is possible to resolve some of these ambiguities using a heuristic, we release the dataset in its current format to encourage research in annotation consolidation and evaluation using these annotations.
Although, we use majority agreement on mention-entities as our gold dataset for all evaluations described later, our released dataset contains non-majority annotations to enable additional research in this domain.

\subsection{\TweetNERD{} Data Format}
\label{sec:data-format}

We release \TweetNERD{} in a non-tokenized format. \TweetNERD{} consists of only Tweet Ids and our annotations as suggested by the Public Twitter API\footnote{\url{https://developer.twitter.com/en/docs/twitter-api}}. Each \TweetNERD{} file consists of Tweet ids, start and end offsets, mention phrase, linked entity, and annotator agreement score (see Figure \ref{tab:data_format}). We provide details in Appendix \ref{sec:ner_conversion} on how to convert this format into token label format suitable for training and evaluating NER systems. All mentions are untyped. 

\begin{figure}[!htbp]
    \centering
    \begin{tabular}{rrrrrr}
    \toprule
    Id & Start & End & Mention & Entity & Score\\
    \midrule
    1 & 7 & 14 & Twitter & Q918 & 3\\
    2 & 0 & 5 & Paris & Q90 & 3\\
    3 & 0 & 4 & Anil & AMB. & 2\\
    \bottomrule
    \end{tabular}
    \caption{\textbf{Data Format.} Sample Tweets from Figure \ref{fig:interface} to illustrate the data format.}
    \label{tab:data_format}
\end{figure}

\section{Evaluation on \TweetNERD}
\label{sec:evaluation}

\setlength{\tabcolsep}{2pt}
\begin{table}[!ht]
\caption{Evaluating \TweetNERDOOD{} and \TweetNERDAcademic{} using existing systems.}
\label{tab:all_eval_scores}
\begin{subfigure}[t]{0.5\linewidth}
    \centering
    \begin{tabular}{l|r|r}
    \toprule
         Model & OOD & Academic \\
    \midrule
         Spacy            & 0.377                         & 0.454                         \\
        StanzaNLP          & 0.421                         & 0.503                         \\
        SocialMediaIE    & 0.153 & 0.245 \\
        BERTweet WNUT17 & 0.278 & 0.46  \\
        TwitterNER     & 0.424                         & 0.522                         \\
        AllenNLP         & 0.454                         & 0.552         \\               
    \bottomrule
    \end{tabular}
    \caption{NER \texttt{strong\_mention\_match} F1 scores.}
    \label{tab:ner_scores}
\end{subfigure}
\begin{subfigure}[t]{0.5\linewidth}
    \centering
    \begin{tabular}{l|rr|rr}
    \toprule
     Model & \multicolumn{2}{c}{\texttt{entity\_match}}                       & \multicolumn{2}{c}{\texttt{strong\_all\_match}}                   \\
     \midrule
     & \multicolumn{1}{r}{OOD} & \multicolumn{1}{r}{Academic} & \multicolumn{1}{r}{OOD} & \multicolumn{1}{r}{Academic} \\
    \midrule
    GENRE  & 0.469 & 0.636 & 0.39  & 0.624 \\
    REL    & 0.463 & 0.614 & 0.387 & 0.56  \\
    Lookup & 0.621 & 0.645 & 0.584 & 0.617 \\
    \bottomrule
    \end{tabular}
    \caption{Entity Linking given true spans (EL) F1 scores.}
    \label{tab:linking_scores}
\end{subfigure}
\begin{subfigure}[t]{0.5\linewidth}
    \centering
    \begin{tabular}{l|rr|rr}
    \toprule
     Model & \multicolumn{2}{c}{\texttt{entity\_match}}                       & \multicolumn{2}{c}{\texttt{strong\_all\_match}}                   \\
     \midrule
     & \multicolumn{1}{r}{OOD} & \multicolumn{1}{r}{Academic} & \multicolumn{1}{r}{OOD} & \multicolumn{1}{r}{Academic} \\
    \midrule
     DBpedia & 0.292 & 0.399 & 0.231 & 0.347 \\
NLAI    & 0.522 & 0.568 & 0.313 & 0.494 \\
TAGME   & 0.402 & 0.431 & 0.293 & 0.381 \\
REL    & 0.344 & 0.484 & 0.27  & 0.444 \\
GENRE\tablefootnote{Using GENRE end-to-end entity linking model for Table 5-c and entity disambiguation model for Table 5-b. Evaluation scores are after removing a few Tweets from the gold set for which the GENRE model fails. Not removing these Tweets and simply returning Null for GENRE only makes a difference in the third decimal point.}   & 0.307 & 0.458 & 0.223 & 0.379 \\
     \bottomrule
    \end{tabular}
    \caption{End to End Entity Linking (End2End) F1 scores.}
    \label{tab:end2end_scores}
\end{subfigure}
\end{table}

We use \texttt{neleval}\footnote{\url{https://neleval.readthedocs.io/}} library for evaluating various publicly available systems on \TweetNERD. For our evaluations we always map \texttt{NOT FOUND} and \texttt{AMBIGUOUS} to \texttt{NIL}. We describe the metrics and the evaluation setup below for the three NERD tasks: Named Entity Recognition (NER), Entity Linking with True Spans (EL), and End to End Entity Linking (End2End). 

\paragraph{Metrics} We first describe the main metrics from \texttt{neleval} that are used for evaluation across the three sub-tasks defined above. 
\texttt{strong\_mention\_match} is a micro-averaged evaluation of entity mentions that is used for the NER task. This metric requires a start and end offset to be returned for the mention. For systems that don't provide offsets we infer the offset in the original text by finding the first mention of the identified mention text.
\texttt{strong\_all\_match} is a micro-averaged link evaluation of all mention-entities whereas
\texttt{entity\_match} is a micro-averaged Tweet-level set of entities measure.
For EL and End2End tasks, we use \texttt{strong\_all\_match} and \texttt{entity\_match} as evaluation metrics. \texttt{entity\_match} is more robust to offset mismatches whereas \texttt{strong\_all\_match} requires a strict match.
We report F1 scores for each metric described above. F1 is a harmonic mean of precision and recall. Please see Appendix ~\ref{sec:appendix_metrics} for details. 

\subsection{Performance of Existing Entity Linking Systems.}
\label{sec:baselines}
In this section we benchmark existing systems for NERD tasks. We provide these benchmarks as a baseline on \TweetNERD. We also report numbers on a simple heuristic baseline using exact match lookup and show that it performs well across our datasets. All experiments run on a machine with single NVIDIA A100 GPU and 32 GB RAM. We choose our baselines based on the availability of existing NER, EL, and End2End systems favoring those systems which are widely used in literature or are specifically built for social media or Tweet datasets.

\paragraph{Named Entity Recognition.}
For NER we use StanzaNLP ~\citep{qi2020stanza}, Spacy\footnote{~\url{https://spacy.io/api/entityrecognizer}}, AllenNLP ~\citep{Peters2017SemisupervisedST}, BERTweet \citep{bertweet}\footnote{\url{https://huggingface.co/socialmediaie/bertweet-base_wnut17_ner}} fine-tuned for NER using WNUT17 \citep{derczynski-etal-2017-results}, Twitter NER \citep{mishra-diesner-2016-semi}, and Social Media IE \citep{mishra-mdmt-2019,Mishra2020SIGIR,MishraThesis2020}. We chose these for their popularity and for their relevance for social media data. See more details about the systems in Appendix Section \ref{sec:ner_systems}. We find that TwitterNER and AllenNLP perform the best on both OOD and Academic dataset. We also find that many of the errors of other systems come from incorrect mention start and end offset prediction even when the mention string is correctly identified.

\paragraph{Entity Linking given True Spans (EL).}\label{Diambiguation}
For EL we use GENRE (Generative ENtity REtrieval) \citep{decao2020autoregressive}, REL (Radboud Entity Linker) \citep{vanHulst:2020:REL}\footnote{\url{https://github.com/informagi/REL}}, and Lookup. Lookup is a simple heuristic based system, where given true mentions, we fetch the most likely entity based on popularity defined via mention candidate co-occurrence in Wikipedia.
See details in Appendix Section \ref{sec:appendix_disambiguation}.
We find that Lookup is a strong baseline for both datasets, whereas REL and GENRE come close in performance on Academic subset.

\paragraph{End to End Entity Linking (End2End).}
\label{para:end2end_eval}

For End2End we use GENRE (Generative ENtity REtrieval) \citep{decao2020autoregressive}, REL (Radboud Entity Linker) \citep{vanHulst:2020:REL}, TagMe \citep{TagMe}\footnote{\url{https://github.com/gammaliu/tagme}}, DBPedia Spotlight \citep{isem2013daiber}, Natural Language AI (NLAI) API from Google \footnote{\url{https://cloud.google.com/natural-language}}.
See details in Appendix Section \ref{sec:appendix_endtoend}.
We find that NLAI is a strong baseline for both datasets, whereas REL and GENRE come close in performance on Academic subset. For OOD subset, NLAI is the best performing model.

\section{Limitations}
\label{sec:limitations}
\TweetNERD{} is the largest dataset for NERD tasks on Tweets. However, we highlight a few limitations. First, this is a non-static dataset since some of the Tweets referenced by Tweet IDs in \TweetNERD{} may become inaccessible at a later date. Our inclusion of \TweetNERDAcademic{} may help mitigate this to some extent as Tweets in that subset have survived a longer duration. Second, because of the difficulty of our annotation task the performance ceiling on \TweetNERD{} is limited as highlighted in the inter-annotator agreement section. However, this provides an opportunity to develop systems on such challenging benchmarks. Finally, the offset based format of \TweetNERD{} makes it challenging to be benchmarked by traditional NER systems which often rely on pre-tokenized text. Our suggestion for using \texttt{neleval} may help address that issue but will require systems to return offsets corresponding to the original text in \TweetNERD{} which may be challenging for traditional systems. The \texttt{entity\_match} eval score is tokenization and offset agnostic but is only applicable for the end to end NERD task.

\section{Conclusion}
\label{sec:conclusion}
We described the largest dataset for NERD tasks on Tweets called \TweetNERD{} and performed benchmarking on popular NERD systems on its two subsets \TweetNERDOOD{} and \TweetNERDAcademic. We hope that the release of this large-scale dataset enables research community to revisit and conduct further research into the problem of entity linking on social media. \TweetNERD{} should foster research and development of robust NERD models for social media which exhibit generalization across domains and time periods. \TweetNERD{} is available at: \url{https://doi.org/10.5281/zenodo.6617192} under Creative Commons Attribution 4.0 International (CC BY 4.0) license \citep{mishra_shubhanshu_2022_6617192}. Check out more details at \url{https://github.com/twitter-research/TweetNERD}.

\begin{ack}
We would like to thank Twitter’s Human Computation team, specifically Iuliia Rivera, and Marge Oreta for their efforts in designing and setting up the annotation tasks and training the annotators which was instrumental in generating \TweetNERD{} data. We would also like to extend our gratitude to the annotators who contributed to this task directly.
\end{ack}

\bibliography{anthology,custom}
\bibliographystyle{plainnat}

\section*{Checklist}

The checklist follows the references.  Please
read the checklist guidelines carefully for information on how to answer these
questions.  For each question, change the default \answerTODO{} to \answerYes{},
\answerNo{}, or \answerNA{}.  You are strongly encouraged to include a {\bf
justification to your answer}, either by referencing the appropriate section of
your paper or providing a brief inline description.  For example:
\begin{itemize}
  \item Did you include the license to the code and datasets? \answerYes{See Introduction.}
\end{itemize}
Please do not modify the questions and only use the provided macros for your
answers.  Note that the Checklist section does not count towards the page
limit.  In your paper, please delete this instructions block and only keep the
Checklist section heading above along with the questions/answers below.

\begin{enumerate}

\item For all authors...
\begin{enumerate}
  \item Do the main claims made in the abstract and introduction accurately reflect the paper's contributions and scope?
    \answerYes{}
  \item Did you describe the limitations of your work?
    \answerYes{See limiations section}
  \item Did you discuss any potential negative societal impacts of your work?
    \answerNo{Not applicable}
  \item Have you read the ethics review guidelines and ensured that your paper conforms to them?
    \answerYes{No potential negative societal impacts required}
\end{enumerate}

\item If you are including theoretical results...
\begin{enumerate}
  \item Did you state the full set of assumptions of all theoretical results?
    \answerNA{}
	\item Did you include complete proofs of all theoretical results?
    \answerNA{}
\end{enumerate}

\item If you ran experiments (e.g. for benchmarks)...
\begin{enumerate}
  \item Did you include the code, data, and instructions needed to reproduce the main experimental results (either in the supplemental material or as a URL)?
    \answerYes{We plan to release code at: \url{https://github.com/twitter-research/TweetNERD}}
  \item Did you specify all the training details (e.g., data splits, hyperparameters, how they were chosen)?
    \answerNA{No training done.}
	\item Did you report error bars (e.g., with respect to the random seed after running experiments multiple times)?
    \answerNA{No training done.}
	\item Did you include the total amount of compute and the type of resources used (e.g., type of GPUs, internal cluster, or cloud provider)?
    \answerYes{See section Performance of Existing Entity Linking Systems.}
\end{enumerate}

\item If you are using existing assets (e.g., code, data, models) or curating/releasing new assets...
\begin{enumerate}
  \item If your work uses existing assets, did you cite the creators?
    \answerYes{}
  \item Did you mention the license of the assets?
    \answerNA{We recreated the existing datasets used for our analysis.}
  \item Did you include any new assets either in the supplemental material or as a URL?
    \answerNo{}
  \item Did you discuss whether and how consent was obtained from people whose data you're using/curating?
    \answerNA{Data in public domain}
  \item Did you discuss whether the data you are using/curating contains personally identifiable information or offensive content?
    \answerYes{See section on \TweetNERDAcademic}
\end{enumerate}

\item If you used crowdsourcing or conducted research with human subjects...
\begin{enumerate}
  \item Did you include the full text of instructions given to participants and screenshots, if applicable?
    \answerNA{We had an inhouse team of annotators and no crowdsourcing was used. We include the details of the guildelines for the annotators under Annotation Setup.}
  \item Did you describe any potential participant risks, with links to Institutional Review Board (IRB) approvals, if applicable?
    \answerNA{}
  \item Did you include the estimated hourly wage paid to participants and the total amount spent on participant compensation?
    \answerNA{We have an inhouse team.}
\end{enumerate}

\end{enumerate}

\clearpage
\appendix

\setcounter{table}{0}
\setcounter{figure}{0}
\renewcommand{\thetable}{A\arabic{table}}
\renewcommand{\thefigure}{A\arabic{figure}}

\section{Converting data to BIO format for NER}
\label{sec:ner_conversion}

In order to convert the dataset to NER format we suggest tokenizing Tweet text and utilizing the character offsets to identify mention tokens. E.g. \texttt{just setting up my twttr} with offsets \texttt{19} and \texttt{24}, and DBpedia category as \texttt{Organization}, can be converted to the NER BIO format as follows: \texttt{tokens, starts, ends = tokenize\_with\_offsets("just setting up my twttr")} and then assigning \texttt{O} labels to all tokens outside the phrase start and end offsets and \texttt{B-ORG} and \texttt{I-ORG} label to all tokens within the phrase offsets. This approach works as long as the tokenizer returned offsets correspond to the offset of the phrase in the original text, i.e. tokenization is non-destructive. See example code in listing \ref{lst:ner_conversion}.

\begin{lstlisting}[linewidth=\textwidth,breaklines=true, language=Python, caption=Conversion of offset format to NER BIO format using one choice of tokenization., float=*,label={lst:ner_conversion}]
import bisect
import re

def tokenize_with_offsets(text):
  """Dummy tokenizer.
  Use any tokenizer you want as long it as the same API."""
  tokens, starts, ends = zip(*[
    (m.group(), m.start(), m.end())
    for m in re.finditer(r'\S+', text)
  ])
  return tokens, starts, ends

def get_labels(starts, ends, spans):
  """Convert offsets to sequence labels in BIO format."""
  labels = ["O"]*len(starts)
  spans = sorted(spans)
  for s,e,l in spans:
    li = bisect.bisect_left(starts, s)
    ri = bisect.bisect_left(starts, e)
    ni = len(labels[li:ri])
    labels[li] = f"B-{l}"
    labels[li+1:ri] = [f"I-{l}"]*(ni-1)
  return labels

text = "just setting up my twttr"
(tokens, starts, ends) = tokenize_with_offsets(text)

# tokens = ["just", "setting", "up", "my", "twttr"]
# starts = [0, 5, 13, 16, 19]
# ends = [4, 12, 15, 18, 24]

spans = [(19, 24, "ORG")]
labels = get_labels(starts, ends, spans)

# labels = ["O", "O", "O", "O", "B-ORG"]
\end{lstlisting}

\section{Metrics} ~\label{sec:appendix_metrics}
\begin{table}[H]
\centering
\caption{NERD Metrics}
\label{tab:nerd-metrics}
\small
\begin{tabular}{@{}l|l@{}}
\toprule
Metric                 & Description                                                                                                                                                                                                                                               \\ \midrule
\texttt{strong\_mention\_match} & \begin{tabular}[c]{@{}l@{}}\texttt{strong\_mention\_match} is a micro-averaged evaluation of entity mentions. \\ A system span must match a gold span exactly to be counted as correct.\end{tabular}                                                               \\ \midrule
\texttt{strong\_all\_match}     & \begin{tabular}[c]{@{}l@{}}\texttt{strong\_all\_match} is a micro-averaged link evaluation of all mentions. \\ A mention is counted as correct if is either a link match or a nil match.\\ A correct nil match must have the same span as a gold
nil. For a correct link \\ match a system link must have the same span and KB
identifier as a gold link. 
\end{tabular} \\ \midrule
\texttt{entity\_match}          & \begin{tabular}[c]{@{}l@{}}\texttt{entity\_match} is a micro-averaged tweet-level set-of-titles measure.\\ It is the same as entity match reported by ~\citep{10.1145/2488388.2488411}\end{tabular}                                         \\ \bottomrule

\end{tabular}

\end{table}

\section{Dataset details}

\paragraph{NER types.} See table \ref{tab:hcomp_types}.
\paragraph{Temporal distribution.} See figure \ref{fig:data_temporal_freq}.

\subsection{Academic Dataset Details}
\label{sec:appendix_academic_data_details}

As explained in section \ref{sec:sampling} it is difficult to sample datasets for NERD tasks to ensure high number of Tweets containing diverse set of entities. Hence, we addressed this sampling issue by including a split based on Tweets already annotated for NERD or related tasks in existing academic benchmarks. This ensures high percentage of Tweets with named entities and linked entities.  Please note not all the datasets we include in \TweetNERDAcademic{} exist for NERD task. Some exist for NED, some for NER, and some for entity aspect extraction, and some for generic NLP tasks like part-of-speech tagging. We have included these datasets as they contain high density of entities and hence can warrant inclusion in a diverse entity linking test set.

\paragraph{Tgx \citep{dredze-etal-2016-twitter}} This dataset is for cross domain co-reference resolution (CDCR). It contains Tweets around the 2013 Grammy music awards ceremony, therefore it mostly contains mentions of Grammy and Music Artists from 2013. Only tweets with person names have been annotated. Original spans detected via NER system and then annotators fixed mention detection issues, grouped similar mentions, and linked to English Wikipedia. Each Tweet annotated by two annotators. No information on annotator agreement provided in the paper. Contains person names who do not occur in Wikipedia.

\paragraph{Broad \citep{derczynski-etal-2016-broad}} This is an NER dataset and hence only contains mention detection annotations. Includes Person, Location, and Organization named entities. Annotations provided by experts and also via crowd-sourcing. They allow annotating username mentions as NE. The dataset has high temporal and geographical diversity with Tweets from 2009 to 2014. They find low agreement among crowd (35\% F1) and gold annotations but high recall of named entities. The inter-annotator agreement is high.

\paragraph{Entity Profiling \citep{spina2012corpus}} Original dataset created for Entity level aspect extraction. Annotation process is non-traditional. We include this dataset for its high availability of named entities.

\paragraph{\textbf{NEEL 2016}  \citep{NEEL2016}} Dataset created for the Making Sense of Microposts (\#Microposts2016) Named Entity rEcognition and Linking (NEEL) Challenge. It cosists of NERD annotations. It includes annotation of Hashtags and user mentions. The dev and test set come from two events from December 2015 around the US primary elections and the Star Wars premiere. 

\paragraph{\textbf{NEEL v2}   \citep{yang-chang-2015-mart}} This dataset is a combination of \citep{microposts2014_neel_cano.ea:2014} and \citep{fang-chang-2014-entity}. It includes Tweets annotated for NERD as well as for Information Retrieval (IR) given an entity as a query.

\paragraph{\textbf{\citet{fang-chang-2014-entity}}} Dataset of Tweets from December of 2012 from verified users containing location information. It contains Tweets annotated for NERD as well as IR task. Tweets only annotated for person, organization, location, event, and others NER type. For the IR task the authors take 10 query entities and sample 100 Tweets per query and assess if the Tweet contains a mention of the query entity. Entities come from Freebase \citep{freebase:datadumps} which contains subset of entities of Wikipedia.

\paragraph{\textbf{Twitter NEED}    \citep{locke2009named}} This dataset consists of Tweets annotated using CoNLL-2003 guidelines. The author allows marking of user mention as named entities. Tweets were collected on February 10 and March 15. It contained Tweets from February 10 about economic recession,  Australian Bushfires, and gas explosion in Bozeman, MT on March 15. They found that Topic related Tweets had much higher rate of named entities.

\paragraph{\textbf{Ark POS}     \citep{gimpel-etal-2011-part}} This dataset was created for Part of Speech tagging for Tweets. It contains 6.4 tokens referring to proper nouns which make it likely to contain sufficient named entities and hence a likely candidate to be included for benchmarking NERD systems for Tweets.

\paragraph{\textbf{WSDM2012}  \citep{Meij-WSDM2012}} It includes 20 Tweets each from a set of verified users. 562 Tweets are manually annotated by two annotators. Annotation was done at the Tweet level where relevant entities for a given Tweet were marked. The authors do not provide agreement rates. The annotated entities may or may not be mentioned explicitly in the text.

\paragraph{\textbf{Yodie} \citep{gorrell_using_2015}} It consists of Tweets annotated using DBPedia URI from financial institutions and news outlets and climate change discussions. The dataset period is 2013-2014. Tweets were tagged using Crowdflower interface using 10 NLP researchers with each Tweet tagged by three annotators. 89\% of entities had unanimous agreement. Tweets were annotated for person, organization, and location entities, while linking included the NIL class. 

\section{Evaluation system details}

\subsection{Named Entity Recognition (NER)}\label{sec:ner_systems}

\paragraph{StanzaNLP ~\citep{qi2020stanza}.}
Stanza is a collection of accurate and efficient tools for the linguistic analysis of many human languages based on the Universal Dependencies (UD) formalism and includes named entity recognition as a functionality. For each document stanza outputs entity mentions and their start and end character offsets which can be directly used for neleval evaluation.

\paragraph[]{Spacy\footnote{~\url{https://spacy.io/api/entityrecognizer}.}}

Spacy NLP library provides a transition-based named entity recognition component. The entity recognizer identifies non-overlapping labelled spans of tokens.  The loss function optimizes for whole entity accuracy, which assumes a good inter-annotator agreement on boundary tokens for good performance. Spacy identified mentions are in the desired character offset format and hence can be directly used for evaluation.

\paragraph{AllenNLP ~\citep{Peters2017SemisupervisedST}.}
The AllenNLP named entity recognizer uses a Gated Recurrent Unit (GRU) character encoder as well as a GRU phrase encoder, and it starts with pretrained GloVe vectors for its token embeddings. It was trained on the CoNLL-2003 NER dataset. AllenNLP outputs BIO labels. To extract mentions and their start and end character offsets we first extract the mentions from the BIO labels corresponding to the non-O tokens. We then perform a search for this phrase in the Tweet text to get the start and end offsets. This leads to some edge cases such as if there are two identical mentions correctly identified, we always count only the first match hence over-penalizing the model. On the other hand, if mention identified by the model was the latter one but only the former mention was part of the gold annotation we under-penalize the model.

\paragraph{Twitter NER \citep{mishra-diesner-2016-semi}.} Twitter NER is a conditional random field model trained specifically for Tweets using a combination of rules, gazetteers, and semi-supervised learning. It is a prominent non-neural baseline for NER on Tweets.

\paragraph{Social Media IE \citep{mishra-mdmt-2019}.} SocialMediaIE is a multi-task model trained on a combination of tasks for social media information extraction. It uses a pre-trained language model along with multi-dataset multi-task learning setup and is jointly trained to perform NER, Part-of-Speech tagging, Chunking, and Supersense tagging.

\subsection{Entity Linking given True Spans (EL)}\label{sec:appendix_disambiguation}

Given true entity mentions from human annotated data, we compare linking only performance (also known as entity disambiguation) using \texttt{entity\_match} and \texttt{strong\_all\_match} from neleval. 

\paragraph{GENRE (Generative ENtity REtrieval) \citep{decao2020autoregressive}.} GENRE is a sequence-to-sequence model that links entities by generating their name in an autoregressive fashion. Its architechture is based on transformers and it fine-tunes BART~\citep{lewis2019bart} for generating entity names, which in this case are corresponding Wikipedia article titles. We used the model that was trained on BLINK + AidaYago2. 

\paragraph[]{REL (Radboud Entity Linker) \citep{vanHulst:2020:REL}\footnote{\url{https://github.com/informagi/REL}}.} REL is an open source toolkit for entity linking. It uses a modular architecture with mention detection and entity disambiguation components. We use REL \emph{with} mentions to get \emph{only} entity disambiguation results here.

\paragraph{Lookup} Lookup is a simple heuristic based system. Given true mentions, we fetch the most likely entity based on popularity defined via mention candidate co-occurrence in wikipedia.

\subsection{End to End Entity Linking (End2End)}\label{sec:appendix_endtoend}

To compare end to end entity linking systems we use \texttt{entity\_match} and \texttt{strong\_all\_match} from neleval. Some of the models mentioned here have been introduced in Section~\ref{sec:appendix_disambiguation}

\paragraph{GENRE.} For end-to-end entity linking, a Markup annotation is used to indicate the span boundaries with special tokens, and the decoder decides to generate a mention span, a link to a mention, or continue to generate the input at each generation step. Therefore, the model is capable of both detecting and linking entities.

\paragraph{REL.} We use  REL \emph{without} mentions to get complete End2End linking results in this case.

\paragraph[]{TagMe \citep{TagMe}\footnote{\url{https://github.com/gammaliu/tagme}}.} It is an end to end system and is based on a directory of links, pages and Wikipedia graph. We use TagME to get linking results.

\paragraph{DBPedia Spotlight \citep{isem2013daiber}.} Spotlight first detects mentions in a two step process; in the first step, all possible mention candidates are generated using different methods, and the second step selects the best candidates based on a score which is a linear combination of selected features (such as annotation probability). The linking/disambiguation part uses cosine similarity and a vector representation which is based on a modification of TF-IDF weights. 

\paragraph[]{Natural Language AI (NLAI) \footnote{\url{https://cloud.google.com/natural-language}}.} We use the \texttt{documents:analyzeEntities} endpoint of the API to get the entities in the Tweet. The system is black-box but is likely to use deep neural network based solutions for entity recognition and entity linking. 

\end{document}